# Non-Gaussian Random Generators in Bacteria Foraging Algorithm for Multiobjective Optimization

Ganesan T[1]*, Vasant P[2] and Elamvazuthi I[3]

[1]School of Chemical Engineering, The University of Adelaide, Adelaide, 5005 South Australia, Australia
[2]Department of Fundamental and Applied Sciences, University Technology Petronas, 31750 Tronoh, Perak, Malaysia
[3]Department of Electrical and Electronics Engineering, University Technology Petronas, 31750 Tronoh, Perak, Malaysia

**Abstract**

Random generators or stochastic engines are a key component in the structure of metaheuristic algorithms. This work investigates the effects of non-Gaussian stochastic engines on the performance of metaheuristics when solving a real-world optimization problem. In this work, the bacteria foraging algorithm (BFA) was employed in tandem with four random generators (stochastic engines). The stochastic engines operate using the Weibull distribution, Gamma distribution, Gaussian distribution and a chaotic mechanism. The two non-Gaussian distributions are the Weibull and Gamma distributions. In this work, the approaches developed were implemented on the real-world multi-objective resin bonded sand mould problem. The Pareto frontiers obtained were benchmarked using two metrics; the hyper volume indicator (HVI) and the proposed Average Explorative Rate (AER) metric. Detail discussions from various perspectives on the effects of non-Gaussian random generators in metaheuristics are provided.

## Background

Among the all state-of-the-art approaches for solving highly complex optimization problems, swarm intelligence (SI) stands as one of the most favoured strategies. This is mainly due to its effectiveness in the search process and its efficiency in terms of computational time [1]. Some of the most common SI-based approaches are particle swarm optimization (PSO) [2], cuckoo search (CS) [3], ant colony optimization (ACO) [4] and bacterial foraging algorithm (BFA) [5]. In the past, one of the most popular techniques employed for solving constrained nonlinear optimization problems is PSO [2]. However, in recent times other strategies such as BFA have become attractive for this purpose. BFA's computational performance has been proven to as good as and at times better than other SI's as well evolutionary techniques [6]. Inspired by the natural behaviour of the *E.Coli* bacterium searching for nutrients in the environment, Pasino [5] proposed the BFA for solving complex optimization problems. The central premise of this approach is that the bacteria tries to maximize its energy per unit time spent during foraging for nutrients while simultaneously evading noxious substances.

Over the past years, many research works has been centred on resin bonded sand mould systems (cold box moulding process) due to its adaptive capabilities in a dynamics environment, its compliance to environmental factors and its high casting quality [7]. This approach is extremely power-saving when it comes to large-scale or mass production [8]. In most cases, resin bonded sand mould systems have excellent flow behaviour. Nevertheless some introduction of vibration and compaction during the moulding process is required. Casting properties are very much influenced by the mould properties (which are in effect dependent on the process parameters during mould development process) [9]. In this work, the multiobjective (MO) optimization model employed for the resin bonded sand system was developed in Surekha et al. [10]. This model was based on a resin bonded sand system where phenol formaldehyde was used as binder and tetrahydrophtalic anhydride was used as a hardener [10]. Since in this work, the problem formulation is MO in nature, the computational technique (BFA) is incorporated with the weighted-sum framework. Using this approach, multiple solutions are obtained for various weights which are then utilized to construct the Pareto frontier.

Most metaheuristic algorithms are endowed with a stochastic engine that generates random values which diversifies the algorithm's search space. These stochastic engines also provide the algorithm with a 'warm-start' prior to the searching process. Therefore, the type of probability distribution function (PDF) that generates the random value in the stochastic engine heavily influences the algorithm's search capability. In most metaheuristic algorithms, the stochastic engine produces random numbers following a Gaussian PDF [11,12]. In other cases, researchers have employed other approaches to enhance the stochastic engine by using chaos-based functions to couple with the Gaussian PDF [13]. These approaches avoid the algorithm from getting stuck in the local optima which causes premature algorithmic convergence [14]. Thus, stochastic engines play a crucial role in the implementation of metaheuristics. Besides Gaussian stochastic engines, another distribution that has rarely been investigated with respect to metaheuristics are the non-Gaussian distributions. These are usually heavy-tailed distributions, for instance distributors [15-18]. It has been known that many real-world systems (engineering, chemical or economic systems) do not behave in a stochastically Gaussian manner but are prone to contain non-Gaussian fluctuations. Thus the primary aim of this work is to analyse the effects of non-Gaussian stochastic engines on the performance of metaheuristics when implemented in a real-world engineering problem. In this work, the influence of the three types of stochastic engines (which are chaos-based, Gaussian-type PDF and Non-Gaussian-type PDF [19]) on the multi objective optimization of the resin-bonded sand mould system. These approaches are compared and discussed in detail. These procedures are executed by the implementation of the BFA technique.

## Model Characteristics

In Surekha et al. [10] the model describing the responses and the outputs of the optimization process was developed and reported. In that work, the mechanical properties of the quartz-based resin bonded sand core system was modelled using Mamdani-based fuzzy logic [20] and genetic algorithm [21] approaches. The multiobjective representation of the optimization model which consists of four objectives as developed in Surekha et al. [10] is as follows:

Maximize $\rightarrow$ Permeability, $f_1$

Maximize $\rightarrow$ Compression Strength, $f_2$

Maximize $\rightarrow$ Tensile Strength, $f_3$

Maximize $\rightarrow$ Shear Strength, $f_4$

subject to *process constraints.* (1)

The response parameters are; $A$, percentage of resin (%), $B$, percentage of hardener (%), $C$, number of strokes and $D$, curing time (minutes). The final formulation of nonlinear regression model developed and the associated constraints are given as follows [10]:

$$f_1 = -333.77 + 614.73A - 27.435B + 630.36C - 18.97D - 168.98A^2 + 0.239B^2 \\ - 76.08C^2 + 0.111D^2 + 2.827AB + 0.575AC + 0.047AD \\ - 0.7701BC + 0.1323BD - 0.1883CD \quad (2)$$

$$f_2 = 2765.36 + 877.869A - 112.778B - 731.934C + 17.9222D - 357.829A^2 \\ + 0.983456B^2 + 52.2310C^2 - 0.0276946D^2 + 14.6571AB + 96.8495AC \\ - 3.74068AD + 7.62554BC - 0.096084BD - 1.27093CD \quad (3)$$

$$f_3 = -354.406 + 211.418A + 17.3611B + 96.7916C + 2.78503D - 44.7516A^2 \\ - 0.173996B^2 - 10.6696C^2 - 0.026223D^2 - 2.08868AB + 6.05542AC \\ + 0.197646AD + 2.07847BC - 0.078904BD + 1.18561CD \quad (4)$$

$$f_4 = 318.163 + 726.696A + 33.3432B - 721.381C + 2.40622D - 210.057A^2 \\ - 0.189623B^2 + 80.1788C^2 + 0.000987D^2 - 1.89739AB + 49.8702AC \\ - 0.32471AD - 1.70998BC - 0.07323BD + 0.306223CD \quad (5)$$

$$A \in [1.5, 2.5], B \in [30, 50], C \in [3, 5], D \in [60, 100] \quad (6)$$

The algorithms employed in this work were programmed using the C++ programming language on a personal computer with an Intel® Core ™ i5 processor running at 3.2 GHz.

## Stochastic Engines

### Gaussian distribution

A random variable, $x \in X$ which is distributed with a mean, $\mu$ and variance, $\sigma^2$ is said to be a Gaussian or normally distributed when the PDF is as follows:

$$G_X(x) = \frac{1}{\sigma\sqrt{2\pi}} \exp\left[-\frac{1}{2}\left(\frac{x-\mu}{\sigma}\right)^2\right] \quad (7)$$

The Gaussian distribution is very general and widely applicable in various fields of studies for modelling real-valued random numbers (e.g., Brownian motion [22] and Monte Carlo simulations [23]). In this work, the standard normal distribution with $\mu=0$ and $\sigma=1$ is employed in the stochastic engine to generate random values in the metaheuristic.

### Weibull distribution

The Weibull distribution is a type of non-Gaussian distribution which is widely implemented in extreme value statistics. Two-parameter Weibull distributions function for a random variable $x \in X$ is defined as follows:

$$W(x) = 1 - \exp\left[-\left(\frac{x}{\lambda}\right)^k\right] \text{ for } x \in R(0, +\infty) \quad (8)$$

where $W(x)$ is the Weibull distribution, $k$ is the shape parameter and $\lambda$ is the scale parameter. It is required that the scale and shape parameter are positive for the Weibull distribution ($k > 0$ and $\lambda > 0$). For $\lambda = 1$, the Weibull distribution takes the form of the exponential distribution. It should be noted that the Weibull distributions around $\lambda$ gets smaller as the value of $k$ increases. In this work, shape and scale parameters are set such that $\lambda = 1$ and $k = 1$. The Weibull distribution has been widely used in areas such as microbiology [24], information systems [25] and meteorology [26].

### Gamma distribution

Similar to the Weibull distribution, the Gamma distribution is another type of non-Gaussian distribution. The Gamma distribution is influenced by its shape, $\alpha$ and rate, $\beta$ parameters. The Gamma distribution, $\gamma(x)$ for a random variable $xX$ is given as follows:

$$\gamma(x) = 1 - \sum_{i=0}^{\alpha-1} \frac{(\beta x)^i}{i!} \exp(-\beta x) \text{ for } \forall i \in Z \text{ such that } \alpha > 0 \text{ and } \beta > 0 \quad (9)$$

The PDF of the Gamma distribution becomes near-symmetrical if there is an increment in the shape factor and the mean as the skewness decreases. As the standard deviation of the distribution increases, the PDF gradually skews to the left and becomes heavy-tailed. The Gamma distribution has been used successfully in climatology, insurance claim models and risk analysis [27-29].

### Chaotic generator

In this work, a one-dimensional chaotic map was used to initialize population of solutions by embedding the map into the random number generation component in the algorithm. The one-dimensional chaotic map, $\psi_n$ is represented as the following:

$$\psi_{n+1} = f(\psi_n) \quad (10)$$

The most widely studied one-dimensional map is the logistic map [15] which is as the following:

$$f(\psi_n) = r_n \psi_n (1 - \psi_n) \quad (11)$$

$$r_{n+1} = r_n + 0.01 \quad (12)$$

where $r_n \in [0, 5]$ and $r_n \in [0, 5]$. In this mapping like all chaotic maps, the dynamics of the system varies for different sets of initial conditions ($\Psi_0$ and $r_o$).

## Bacteria Foraging Algorithm

In the BFA, four main levels of loops are present in the technique (chemotaxis, swarming, reproduction and elimination-dispersal loops). These loops manage the main functional capabilities of the BFA. Each of the mentioned loops are designed according to bacteria foraging strategies and principles from evolutionary biology. These loops are executed iteratively until the total number of iterations, $N_T$ is satisfied.

Each of the main loops may be iterated until some fitness condition is satisfied or until a user-defined loop cycle limit (chemotaxis ($N_c$), swarming ($N_s$), reproduction ($N_r$) and elimination-dispersal ($N_{ed}$)) is reached.

In chemotaxis, the bacteria with the use of its flagellum, swims and tumbles towards the nutrient source. The tumbling mode allows bacterium motion in a fixed direction while the tumbling mode enables the bacterium to augment its search direction accordingly. Applied in tandem, these two modes give the bacterium capability to stochastically move towards a sufficient source of nutrient. Thus, computationally chemotaxis is presented as follows:

$$\theta^i(j+1,k,l,m) = \theta^i(j,k,l,m) + C(i)\frac{\Delta(i)}{\sqrt{\Delta(i)\Delta^T(i)}} \qquad (13)$$

where $\theta^j(j+1,k,l,m)$ is the $i^{th}$ bacterium at the $j^{th}$ chemotactic step, $k^{th}$ swarming step and $l^{th}$ reproductive step and $m^{th}$ elimination-dispersal step. $C(i)$ is the size of the step taken in a random direction which is fixed by the tumble, and $\Delta \in [-1,1]$ is the random vector.

In the swarming phase, the bacterium communicates to the entire swarm regarding the nutrient profile it mapped during its movement. The communication method adopted by the bacterium is cell-to-cell signalling. In *E. Coli* bacteria, aspartate is released by the cells if it is exposed to high amounts of succinate. This causes the bacteria to conglomerate into groups and hence move in a swarm of high bacterial density. The swarming phase is mathematically presented as follows:

$$J(\theta, P(j,k,l,m)) = \sum_{i=1}^{S}\left[-D_{att}\exp\left(-W_{att}\sum_{m=1}^{P}(\theta_m - \theta_m^i)^2\right)\right] + \sum_{i=1}^{S}\left[-H_{rep}\exp\left(-W_{rep}\sum_{m=1}^{P}(\theta_m - \theta_m^i)^2\right)\right] \qquad (14)$$

where $J(\theta, P(j,k,l,m))$ is the computed dynamic objective function value (not the real objective function in the problem), $S$ is the total number of bacteria, $P$ is the number of variables to be optimized (embedded in each bacterium), $H_{rep}$, $W_{rep}$, $H_{att}$, and $W_{att}$ are user-defined parameters.

During reproduction, the healthy bacteria or the bacteria which are successful in securing a high degree of nutrients are let to reproduce asexually by splitting into two. Bacteria which do not manage to perform according to the specified criteria are eliminated from the group and thus not allowed to reproduce causing their genetic propagation (in this case their foraging strategies) to come to a halt. Due to this cycle, the amount of individual bacterium in the swarm remains constant throughout the execution of the BFA.

Catastrophic events in an environment (such as a sudden change in physical/chemical properties or rapid decrease in nutrient content) can effect in death to a population of bacteria. Such events can cause bacteria to be killed and some to be randomly dispersed to different locations in the objective space. These events which are set to occur in the elimination/dispersal phase help to maintain swarm diversity to make sure the search operation is efficient. The pseudo code for the BFA algorithm employed in this work is provided below:

**START PROGRAM**

Initialize all input parameters ($P$, $H_{rep}$, $W_{rep}$, $H_{att}$, $W_{att}$, $N_T$, $N_c$, $N_r$, $N_s$, $N_{ed}$)

***STOCHASTIC GENERATOR*** - Generate a randomly located swarm of bacteria throughout the objective space

Evaluate bacteria fitness in the objective space

**For** $i=1 \rightarrow N_T$ **do**

**For** $l=1 \rightarrow N_r$ **do**

**For** $m=1 \rightarrow N_{ed}$ **do**

**For** $j=1 \rightarrow N_c$ **do**

**For** $k=1 \rightarrow N_s$ **do**

Perform *chemotaxis* – bacterium swim and tumble until maximum fitness/loop cycle limit is reached

Perform *swarming* – bacterium swarm until maximum fitness/loop cycle limit is reached

**End For**

**End For**

**If** bacterium healthy/maximally fit **then** split and *reproduce*

**Else** *eliminate* remaining bacterium

**End For**

Execute catastrophic *elimination* by assigning some probability of elimination to the swarm. Similarly *disperse* the remaining swarm randomly.

**End For**

**End For**

**END PROGRAM**

The parameter settings specified in all the BFA variants employed in this work is shown in Table 1.

## Measurement Methods

### Hypervolume indicator

The Hyper volume Indicator (HVI) is a strictly Pareto-compliant indicator that is used to measure the quality of solution sets in MO optimization problems [30,31]. Strictly Pareto-compliant can be defined such that if there exists two solution sets to a particular MO problem, then the solution set that dominates the other would a higher indicator value. The HVI measures the volume of the dominated section of the objective space and can be applied for multi-dimensional scenarios. When using the HVI, a reference point needs to be defined. Relative to this point, the volume of the space of all dominated solutions can be measured. The HVI of a solution set $x_d \in X$ can be defined as follows:

$$HVI(X) = vol\left(\bigcup_{(x_1,...x_d) \in X}[r_1,x_1]\times...\times[r_d,x_d]\right) \qquad (15)$$

| Parameters | Values |
|---|---|
| Total Iteration, $N_T$ | 200 |
| Population Size, $P$ | 25 |
| Swimming Loop limit, $N_S$ | 5 |
| Repellent Signal Width, $W_{rep}$ | 10 |
| Attractant Signal Width, $W_{att}$ | 0.2 |
| Repellent Signal Height, $H_{rep}$ | 0.1 |
| Attractant Signal Height, $H_{att}$ | 0.1 |
| Reproduction limit, $N_r$ | 5 |
| Elimination limit, $N_e$ | 5 |

**Table 1:** The individual solutions for the G-BFA approach.

where $r_1,...,r_d$ is the reference point and $vol(.)$ being the usual Lebesgue measure. In this work the HVI is used to measure the quality of the approximation of the Pareto front by the GSA and the DE algorithms when used in conjunction with the weighted sum approach.

**Average explorative rate**

A novel metric, the Average Explorative Rate (AER) is introduced in this work for the purpose of measuring the thoroughness of the search operation carried out by the computational technique in the regions of the objective space. The AER performs online measurements successively during the execution of the computational technique. This metric measures the amount of search region on average covered by the computational technique at each iteration. The proposed AER can be computed by first determining the deviation of the objective function values at each iteration:

$$\delta_n = \left[\frac{f^{n+1}(x_i) - f^n(x_i)}{f^n(x_i)}\right] \quad (16)$$

where $f^n(x_i)$ is the objective function value at the $n^{th}$ iteration with $x_i$ is the decision variables. Then the Heaviside Step Function is employed to return a value if the deviation, $\delta$ is more than some pre-defined value, $L$.

$$H(\delta_n) = \begin{cases} 0 & \text{if } \delta_n < L \\ 1 & \text{if } \delta_n \geq L \end{cases} \quad (17)$$

where $H(\delta_n)$ is the Heaviside Step Function. The AER ($E_R$) is then computed as follows:

$$E_R = \sum_{n=1}^{N}\left(\frac{H(\delta_n)}{N}\right) \quad (18)$$

where $n$ is the iteration count and $N$ is the maximum number of iteration. Therefore the larger the AER value, the more objective space is covered by the computational technique per iteration. This in effect results in a better search operation. It should be noted that when comparing computational techniques, the threshold value, $L$ must be consistent throughout the computational experiments.

**Computational Results and Analysis**

The BFA technique is executed with four of the stochastic engines discussed previously. The BFA technique equipped with the Gaussian, Weibull and Gamma distributions are termed G-BFA, W-BFA and $\gamma$-BFA respectively while the BFA coupled with the chaotic generator is called the Ch-BFA. The solution sets generated by the BFA variants are employed to construct the approximate Pareto frontier. For the approximation of the Pareto frontier, 53 solutions for various weights were obtained for each of the BFA variants employed in this work. The quality of these solution sets was measured using the HVI. The nadir point employed as a reference in the HVI is ($q_1$, $q_2$, $q_3$, $q_4$) = (0, 0, 0, 0). In this work, since the approaches are stochastic in nature, each solution point is selected by taking the best solution obtained from 10 independent runs of the algorithms (for each of the individual weights). The individual solutions for specific weights of the BFA variants were gauged based on the value of the aggregate objective function (weighted-sum approach) given as follows:

$$F = \sum_{\forall i \in [1,4]} w_i f_i \quad (19)$$

where $F$ is the aggregate objective function, $f_i$ are the individual objective functions and $w_i$ are the respective weights. This way the best, median and worst solution was determined. The individual solutions for the G-BFA approach and their aggregate objective values are as in Table 2.

The associated weights ($w_1$, $w_2$, $w_3$, $w_4$) for the best, median and worst solution are (0.1, 0.7, 0.1, 0.1), (0.5, 0.1, 0.3, 0.1) and (0.1, 0.1, 0.7, 0.1). The approximate Pareto frontier obtained using the G-BFA approach is shown in Figure 1. The individual solutions for the W-BFA approach and their aggregate objective values are as in Table 3 and Figure 2 provides depiction of the approximate Pareto frontier obtained using the W-BFA approach. The associated weights ($w_1$, $w_2$, $w_3$, $w_4$) for the best, median and worst solution provided by the W-BFA approach (refer to Table 2) are (0.1, 0.7, 0.1, 0.1), (0.5, 0.1, 0.2, 0.2) and (0.1, 0.1, 0.7, 0.1). The individual solutions for the $\gamma$-BFA approach and their respective values of the aggregate objective functions are shown in Table 4. The associated weights ($w_1$, $w_2$, $w_3$, $w_4$) for the best, median and worst solutions produced by $\gamma$-BFA approach are (0.7, 0.1, 0.1, 0.1), (0.1, 0.4, 0.2, 0.3) and (0.1, 0.1, 0.6, 0.2). The approximate Pareto frontier obtained using the $\gamma$-BFA approach is shown in Figure 3.

The individual solutions for the Ch-BFA variant and their aggregate objective values are as in Table 5 and Figure 4 shows the approximate Pareto frontier obtained using the Ch-BFA approach.

The associated weights ($w_1$, $w_2$, $w_3$, $w_4$) for the best, median and worst solutions produced by Ch-BFA approach are (0.1, 0.7, 0.1, 0.1), (0.1, 0.4, 0.3, 0.2) and (0.1, 0.1, 0.7, 0.1). The degree of dominance for the entire Pareto frontiers produced by all four BFA variants is presented in Figure 5.

In Figure 5, it can be observed that the W-BFA generates the most dominant Pareto frontier followed by the G-BFA, Ch-BFA and $\gamma$-BFA respectively. The frontier obtained using the W-BFA approach is more dominant than G-BFA, Ch-BFA and $\gamma$-BFA by 6.156%, 38.385% and 71.394% respectively. It can be observed in Figure 2 that the frontier produced by the W-BFA is localized at the most optimal regions in the objective space. The G-BFA and the Ch-BFA generates solution set which are highly localized in certain regions (Figures 1 and 4). Thus, these approaches have very limited solution coverage which thus affects the overall dominance of the Pareto frontier. The $\gamma$-BFA on the other hand has a wide spread of solutions and hence high area of coverage (Figure 3). However, the solutions produced by the $\gamma$-BFA are not located on the optimal or dominant regions of the objective space. It can be seen in this work that wide coverage of solutions on the Pareto frontier is a critical criteria for Pareto dominance. Nevertheless, if the solutions are not located in the dominant/optimal regions in the objective space, the overall frontier may not be the highly dominant albeit widely spread. Throughout these executions, the Ch-BFA does not perform well as compared to the (Table 5 and Figure 5) $\gamma$-BFA and the W-BFA. Thus, it is clear that for this problem the chaotic component in the stochastic engine does not improve the performance of the metaheuristic approach. Each of the solutions generated BFA variants are ranked based on the aggregate objective function value as in Tables 2-5. Table 6 shows the best individual solution produced by each of the BFA variants. Similar to the degree of dominance of the entire Pareto frontier, the W-BFA produces outranks the all the other BFA variants in terms of best individual solution. The W-BFA outranks the $\gamma$-BFA, G-BFA and Ch-BFA by 32.857%, 8.538% and 22.113% respectively. From Table 6 and Figure 5, the dominance rankings of the Pareto frontier and the individual solution rankings produced by the BFA variants are synchronous. The computational time of the approaches employed in this work was not compared. This is because all the approaches employed here are BFA-based and they only vary in terms of their stochastic engines which do not significantly contribute to their computational complexity. Therefore comparative analysis related to computational complexity and computational time of the approaches is not investigated in this work. The proposed AER metric was utilized to measure the rate of exploration carried out by each of the approaches. The values of the obtained AER are shown in Figure 6.

| Description | | Best | Median | Worst |
|---|---|---|---|---|
| **Objective Function** | $f_1$ | 841.718 | 842.443 | 876.724 |
| | $f_2$ | 973.687 | 974.652 | 1012.8 |
| | $f_3$ | 312.121 | 312.627 | 333.673 |
| | $f_4$ | 424.551 | 424.027 | 392.404 |
| **Decision Variable** | A | 2.25034 | 2.24784 | 2.25402 |
| | B | 31.2589 | 31.2535 | 31.2635 |
| | C | 4.76753 | 4.75909 | 4.77303 |
| | D | 62.2761 | 62.2647 | 62.2825 |
| **Aggregated function** | F | 839.42 | 654.878 | 461.764 |

**Table 2:** The individual solutions for the G-BFA approach.

| Description | | Best | Median | Worst |
|---|---|---|---|---|
| **Objective Function** | $f_1$ | 763.173 | 763.105 | 763.539 |
| | $f_2$ | 1082.75 | 1082.95 | 1081.66 |
| | $f_3$ | 329.961 | 329.88 | 330.403 |
| | $f_4$ | 438.523 | 438.221 | 440.174 |
| **Decision Variable** | A | 2.49418 | 2.49545 | 2.48713 |
| | B | 37.4942 | 37.4954 | 37.4871 |
| | C | 4.7164 | 4.71767 | 4.70935 |
| | D | 89.7164 | 89.7177 | 89.7094 |
| **Aggregated function** | F | 911.09 | 643.467 | 459.819 |

**Table 3:** The individual solutions for the W-BFA approach.

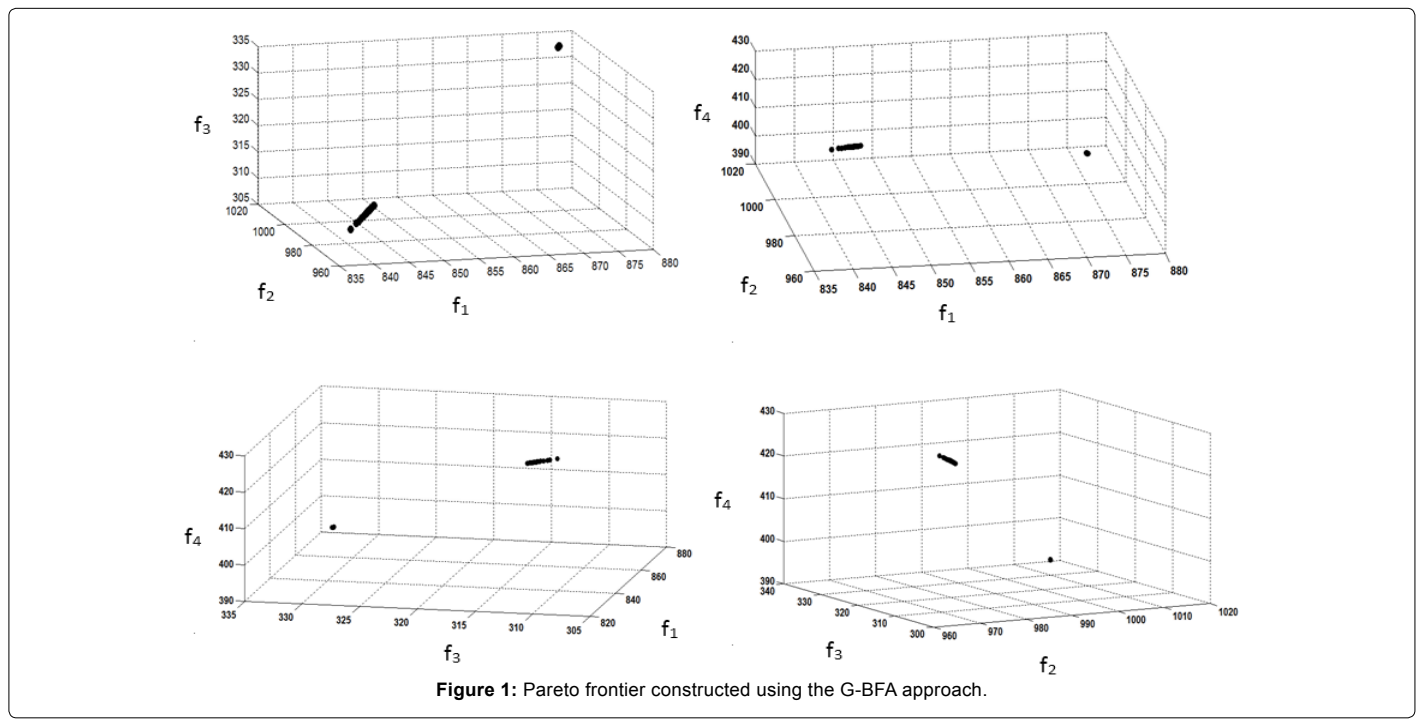

**Figure 1:** Pareto frontier constructed using the G-BFA approach.

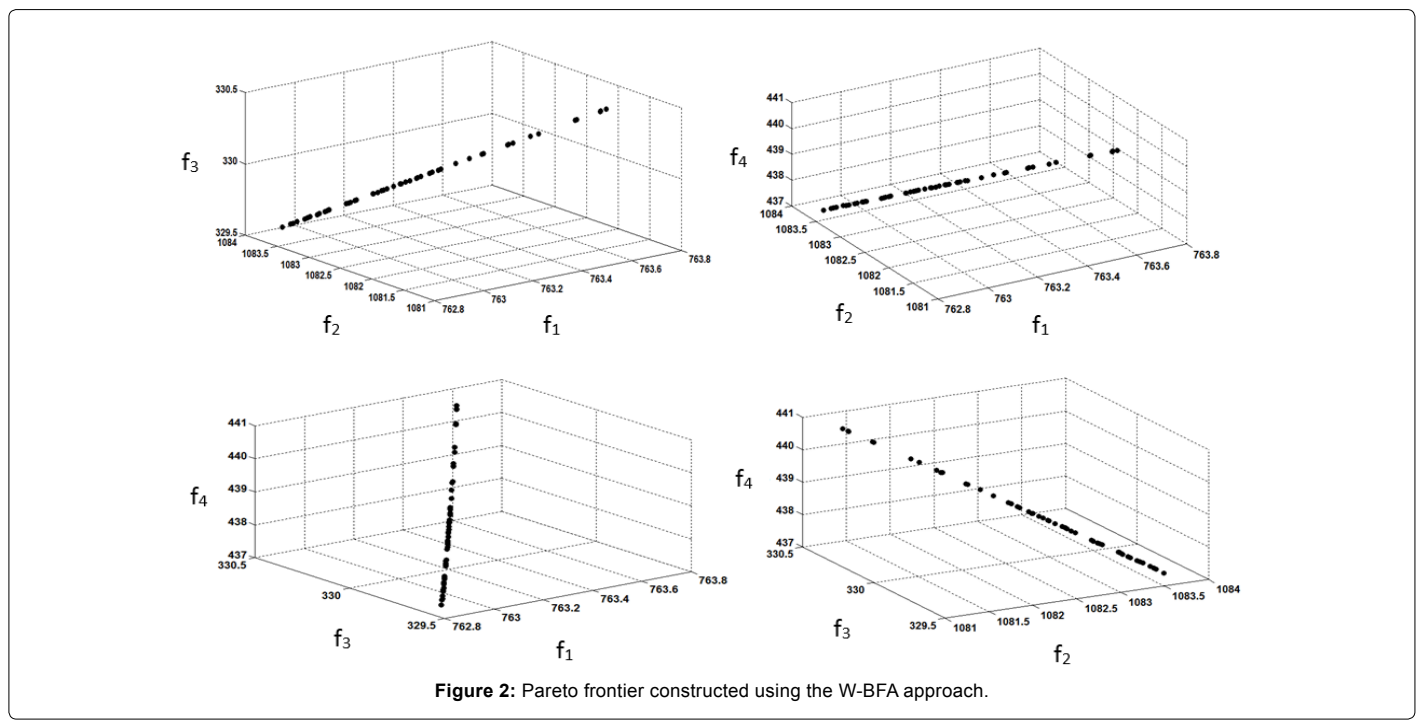

**Figure 2:** Pareto frontier constructed using the W-BFA approach.

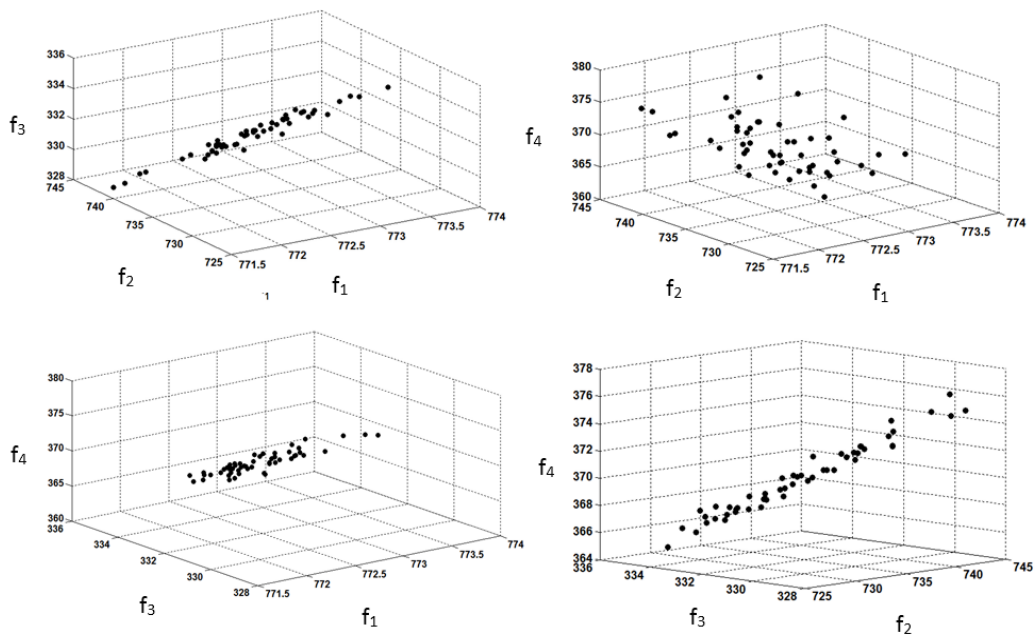

Figure 3: Pareto frontier constructed using the $\gamma$-BFA approach.

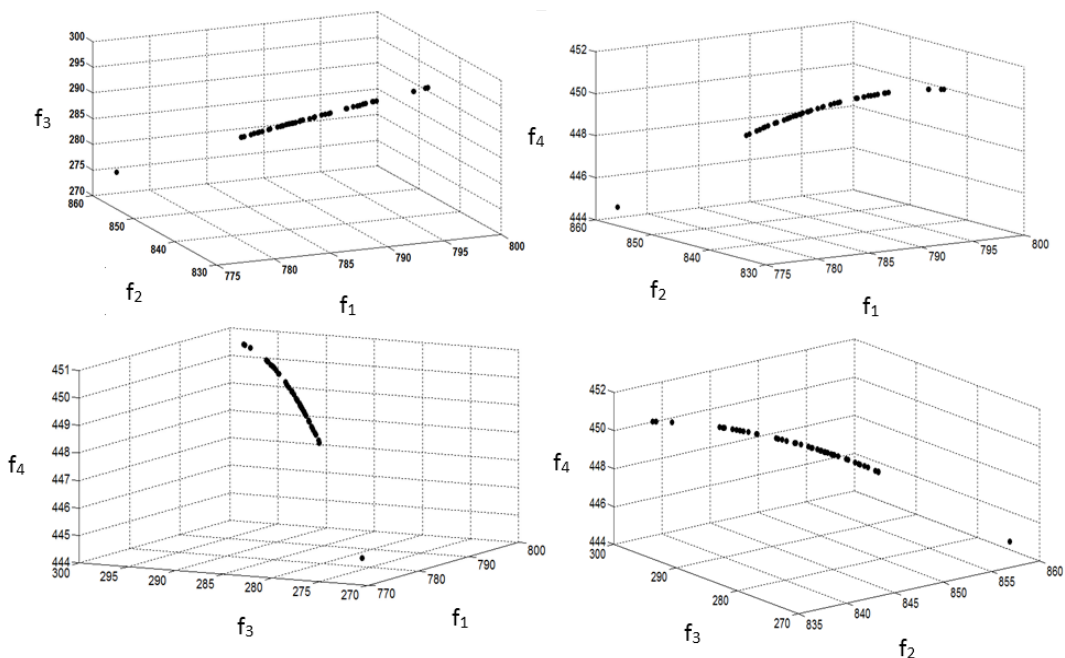

Figure 4: Pareto frontier constructed using the Ch-BFA approach.

| Description | | Best | Median | Worst |
|---|---|---|---|---|
| Objective Function | $f_1$ | 772.963 | 771.625 | 772.975 |
| | $f_2$ | 741.781 | 741.492 | 733.283 |
| | $f_3$ | 328.932 | 328.208 | 333.422 |
| | $f_4$ | 376.205 | 375.26 | 368.473 |
| Decision Variable | A | 1.58506 | 1.57975 | 1.56785 |
| | B | 30.1415 | 30.1386 | 30.1109 |
| | C | 3.19791 | 3.19749 | 3.15392 |
| | D | 60.2543 | 60.2564 | 60.197 |
| Aggregated function | F | 685.766 | 551.979 | 424.374 |

**Table 4:** The individual solutions for the γ-BFA approach.

| Description | | Best | Median | Worst |
|---|---|---|---|---|
| Objective Function | $f_1$ | 787.754 | 787.201 | 788.302 |
| | $f_2$ | 848.311 | 848.971 | 847.645 |
| | $f_3$ | 286.073 | 285.43 | 286.716 |
| | $f_4$ | 449.033 | 448.853 | 449.205 |
| Decision Variable | A | 2.26923 | 2.27358 | 2.26485 |
| | B | 30.7692 | 30.7736 | 30.7649 |
| | C | 3.76923 | 3.77358 | 3.76485 |
| | D | 60.7692 | 60.7736 | 60.7649 |
| Aggregated function | F | 746.103 | 593.708 | 409.216 |

**Table 5:** The individual solutions for the Ch-BFA approach.

| Description | | W-BFA | Y-BFA | G-BFA | Ch-BFA |
|---|---|---|---|---|---|
| Objective Function | $f_1$ | 763.173 | 772.963 | 841.718 | 787.754 |
| | $f_2$ | 1082.75 | 741.781 | 973.687 | 848.311 |
| | $f_3$ | 329.961 | 328.932 | 312.121 | 286.073 |
| | $f_4$ | 438.523 | 376.205 | 424.551 | 449.033 |
| Decision Variable | A | 2.49418 | 1.58506 | 2.25034 | 2.26923 |
| | B | 37.4942 | 30.1415 | 31.2589 | 30.7692 |
| | C | 4.7164 | 3.19791 | 4.76753 | 3.76923 |
| | D | 89.7164 | 60.2543 | 62.2761 | 60.7692 |
| Aggregated function | F | 911.09 | 685.766 | 839.42 | 746.103 |

**Table 6:** The best individual solutions produced by the BFA variants.

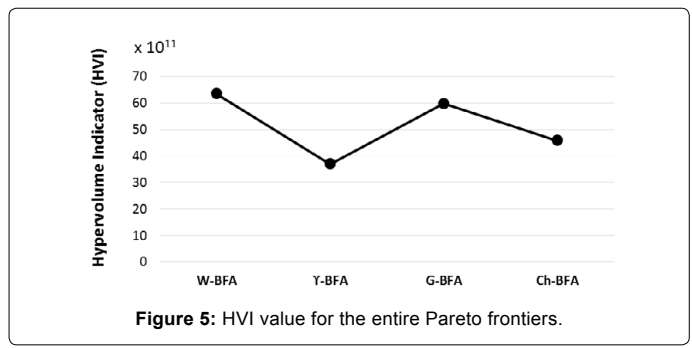

**Figure 5:** HVI value for the entire Pareto frontiers.

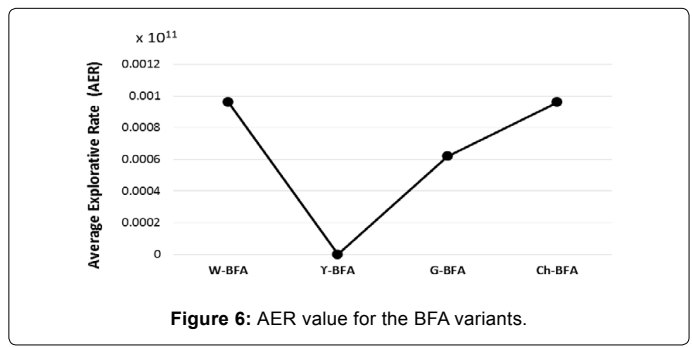

**Figure 6:** AER value for the BFA variants.

In Figure 6, the AER values are observed to be at maximal for the W-BFA approach followed by the Ch-BFA and the G-BFA. The lowest AER value was produced by the -BFA. It can be seen the AER value is related to the degree of dominance determined by the HVI (Figure 5). The W-BFA produces the most dominant Pareto frontier and the highest AER value. Similarly the Pareto frontier generated by the γ -BFA has the lowest dominance and the lowest AER value while Ch-BFA and the G-BFA fall into the middle rank when measured with both the metrics. Thus, the explorative rate clearly indicates the effectiveness of the approach in terms of obtaining optimal solutions which construct dominant Pareto frontiers. The AER can thus be employed as an online metric to gauge and increase the adaptability of the approach during execution.

In this work, the weighted-sum scalarization approach was employed to construct the Pareto frontier from solutions with various weightages. Although the weighted sum approach is very apt for Pareto frontier construction in multi-objective scenarios, this approach fails to guarantee Pareto optimality [32]. Scalarization techniques such as the weighted-sum approach are incapable to accurately approximate Pareto frontiers which are concavely shaped. Since this problem is a maximization problem, the nadir point was chosen such that all the solution points obtained dominate this point. This makes the benchmarking results obtained using the HVI independent of the choice of the nadir point.

All the BFA-based computational approaches employed in this work performed stable computations during program executions. All solution points obtained using the approaches were feasible and no constraint violations occurred. One of the advantages of the BFA approach is that it performs a very thorough search since it has many cascaded loops (chemotaxis, swarming, reproduction and elimination-dispersal. Since this increases the computational complexity of the BFA approach, the negative impact on the execution time is inevitable.

## Outlook

The introduction of the non-Gaussian approach in the conventional stochastic engine has shown very interesting results. It can be observed in Table 5 and Figure 6 that the W-BFA which is equipped with a Non-Gaussian stochastic engine outranks all the other approaches employed in this work in terms of individual solution and degree of Pareto frontier dominance. However, the γ -BFA does not perform as well the W-BFA or the conventional G-BFA although it is equipped with a non-Gaussian stochastic engine as well. These results show that although having a non-Gaussian stochastic engine may be advantageous, it is possible that the choice of non-Gaussian distribution employed in the solution method may be dependent on the type of solution landscape. Due to this dependence, the effectiveness of the computational approach with a particular type of non-Gaussian stochastic engine would vary based on the problem characteristics and type.

The proposed AER metric was observed to provide good correlation with the measurements employed in this work. In future works, the AER metric could be employed in more computational approaches as an effective tool to direct the search effectively during program execution. In addition, other types of non-Gaussian distributions such the Gumbel distribution [15] could be tested as a stochastic engine in other types of metaheuristics (e.g., evolutionary algorithms:

Genetic Programming [33] and Differential Evolution [12]). Besides, more real-world multiobjective problems should be explored using computational techniques with non-Gaussian stochastic engines to investigate and validate the effectiveness of this framework in solving optimization problems.